\pdfoutput=1

\documentclass[11pt]{article}

\usepackage[final]{acl}

\usepackage{times}
\usepackage{latexsym}
\usepackage{multirow}
\usepackage{enumitem}
\usepackage{array}
\usepackage{arydshln}
\usepackage{enumitem}

\usepackage[T1]{fontenc}

\usepackage[utf8]{inputenc}
\usepackage{microtype}

\usepackage{inconsolata}

\usepackage{graphicx}

\title{TeXpert: A Multi-Level Benchmark for Evaluating \LaTeX{} Code Generation by LLMs}

\author{
  Sahil Kale\textsuperscript{1}\thanks{Corresponding author. Email: sahil@k-v.ai} \quad
  Vijaykant Nadadur\textsuperscript{1} \\
  \textsuperscript{1}Knowledgeverse AI \\
  \texttt{\{sahil, vrn\}@k-v.ai}
}

\begin{document}
\maketitle
\begin{abstract}
LaTeX's precision and flexibility in typesetting have made it the gold standard for the preparation of scientific documentation. Large Language Models (LLMs) present a promising opportunity for researchers to produce publication-ready material using LaTeX with natural language instructions, yet current benchmarks completely lack evaluation of this ability. By introducing TeXpert, our benchmark dataset with natural language prompts for generating LaTeX code focused on components of scientific documents across multiple difficulty levels, we conduct an in-depth analysis of LLM performance in this regard and identify frequent error types. Our evaluation across open and closed-source LLMs highlights multiple key findings: LLMs excelling on standard benchmarks perform poorly in LaTeX generation with a significant accuracy drop-off as the complexity of tasks increases; open-source models like DeepSeek v3 and DeepSeek Coder strongly rival closed-source counterparts in LaTeX tasks; and formatting and package errors are unexpectedly prevalent, suggesting a lack of diverse LaTeX examples in the training datasets of most LLMs. Our dataset, code, and model evaluations are available on GitHub. \footnote{\url{https://github.com/knowledge-verse-ai/TeXpert}}
\end{abstract}

\section{Introduction}

LaTeX is a highly versatile and widely adopted document preparation system built over the TeX typesetting program \citep{latex_project}. With research-specific advantages including robust handling of mathematical equations, simple formatting commands, and straightforward management of references, it is a popular choice to produce publication-ready scientific material \citep{bos2023latexmetadatapublishingworkflows}.

The recent emergence of LLMs across various applications \citep{garcia-ferrero-etal-2024-medmt5,sherifi2024potential,zhao-etal-2024-optimizing} coupled with improved instruction-following ability \citep{yin-etal-2023-llm,he-etal-2024-complex} prompts an essential research question: "\textit{Can LLMs generate publication-ready LaTeX code for components of scientific documents from natural language instructions?}". Through this research, we aim to evaluate the capability of LLMs in generating syntactically and logically accurate LaTeX code (which we refer to as accurate LaTeX code generation or simply LaTeX generation) and analyse the main types of errors they encounter.

While certain aspects of LaTeX code generation with LLMs, especially for mathematical content \citep{zou2024stempomevaluatinglanguagemodels,zhang2024mathversedoesmultimodalllm}, have been significantly studied, a comprehensive study of LLMs' LaTeX generation ability for various components commonly used in scientific documents (such as tables, figures, bibliography, etc.) remains unexplored. We believe a comprehensive benchmark for evaluating LLMs on LaTeX generation offers two key benefits: analysing common errors LLMs make in generating LaTeX code can provide format and error-based hints for flagging AI-generated research material \citep{chamezopoulos-etal-2024-overview}, and delineating the complexity of LaTeX tasks that LLMs can reliably perform can greatly reduce researchers’ effort on formatting and typesetting.

In this work, we evaluate a diverse range of closed-source and open-source LLMs on their LaTeX generation capabilities. The main contributions of this paper can be stated as follows:
\begin{enumerate}[noitemsep, topsep=0pt]
    \item We introduce TeXpert, a benchmark designed to evaluate LLMs in generating accurate LaTeX code from natural language instructions, focused on commands in scientific documents
    \item We evaluate popular open and closed-source LLMs on TeXpert by computing the success rate across three difficulty classes
    \item We provide comprehensive insights pertaining to LLM limitations in LaTeX generation and identify frequent error types
\end{enumerate}

\section{Related Work}
Existing works on the evaluation of LLMs treat LaTeX-based tasks only as a peripheral component or limit their scope to specific output formats. The ability of LLMs to generate mathematical LaTeX equations from various sources has been explored in datasets like MATH \citep{hendrycks2021measuringmathematicalproblemsolving}, MathBridge \citep{jung2024mathbridgelargecorpusdataset} and STEM-POM \citep{zou2024stempomevaluatinglanguagemodels}. Similarly, the STRUC-BENCH dataset \citep{tang2024strucbenchlargelanguagemodels} contains natural language inputs to test LLMs’ LaTeX generation ability specific only to tabular content. The im2latex-100k dataset \citep{deng2017imagetomarkupgenerationcoarsetofineattention} also focuses on the narrow aspect of testing the ability of LLMs to convert images of mathematical formulae into LaTeX code, while Image2struct \citep{roberts2024image2structbenchmarkingstructureextraction} includes testing vision-language models in extracting structured LaTeX information from images. 

A straightforward idea to evaluate the natural language to LaTeX ability of LLMs would be to generate free-to-use LaTeX templates \footnote{\url{https://www.overleaf.com/latex/templates}} representing various document styles and formats using textual queries. However, these templates are often too large to be directly generated by large language models (LLMs) and are constrained only to a standard set of basic commands, limiting their applicability in this research. Several instruction-following benchmarks for LLMs evaluate their ability to follow natural language commands \citep{qin-etal-2024-infobench,chen-etal-2024-sifo}; however, there is a notable absence of datasets specifically designed to assess models in LaTeX code generation for scientific material.

Identifying and acting upon this need, we present TeXpert, an organised dataset designed to evaluate LLMs' capability to generate syntactically and logically correct LaTeX code from textual descriptions, focused on scientific document components.

\section{Dataset Construction}
\begin{figure}[t]
  \includegraphics[width=\columnwidth]{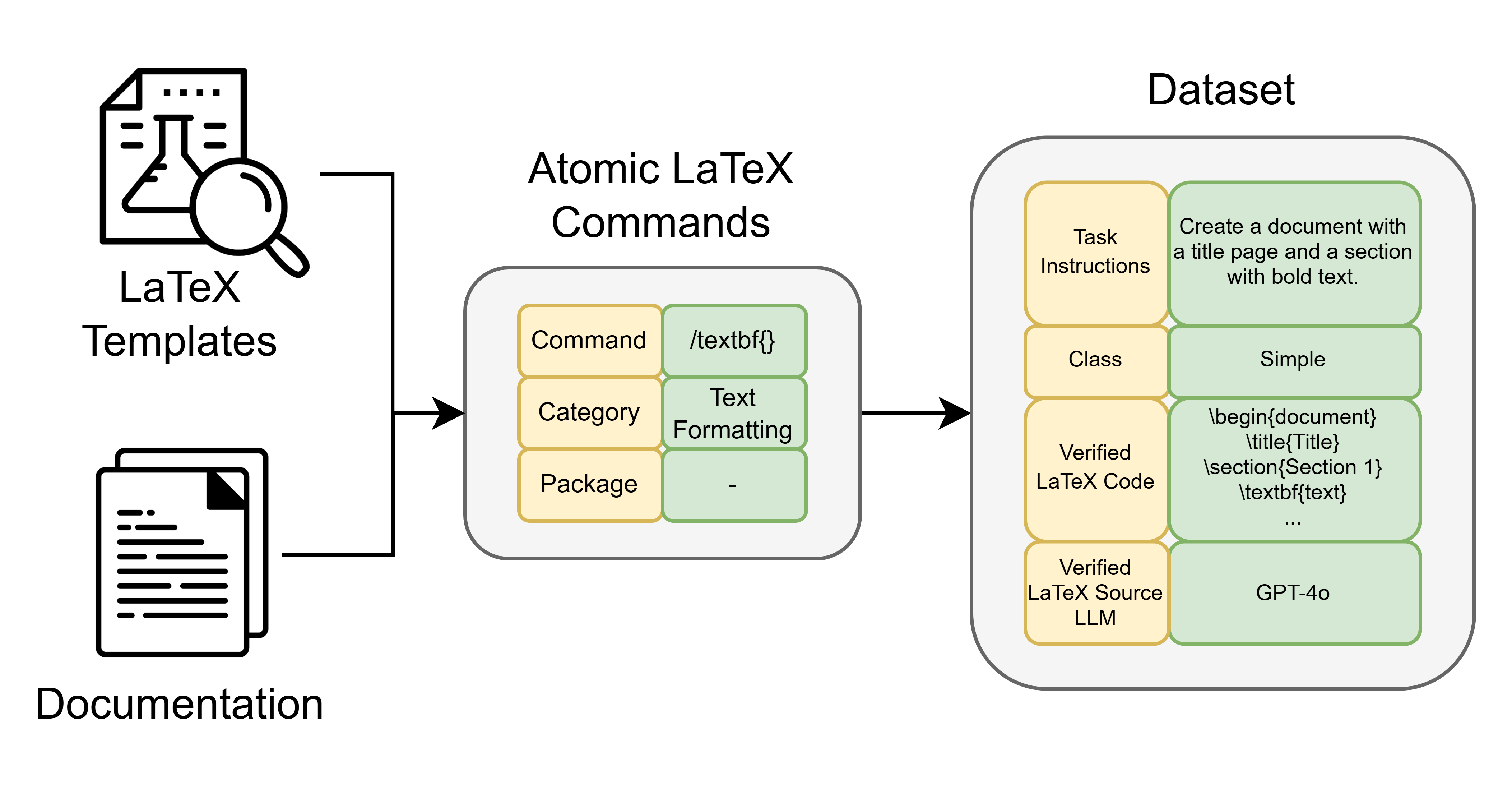}
  \caption{Process used to construct TeXpert, along with the dataset schema}
  \label{fig:process}
\end{figure}

To assess LLMs’ capability to convert unstructured textual descriptions to LaTeX code, we build a benchmark dataset by following the process described in Figure \ref{fig:process}. The process involves two major steps: \\

\noindent\textbf{Collecting atomic LaTeX commands:} We begin by systematically analyzing a range of data sources and scientific document templates to collect atomic LaTeX commands (details of sources and methodology are provided in Appendix \ref{sec:appendix-a1}). These atomic commands, representing the minimal functional units commonly used in scientific writing and typically consisting of a backslash followed by a keyword and optional arguments, were extracted to form the basis of our dataset. The commands were then classified into 5 categories based on their purpose, as shown in Table \ref{tab:atomic}. By adding an extra base step of collecting atomic commands commonly found in scientific formats, we regulate the scope of our final dataset containing LaTeX code generation tasks. 

\begin{table}
\centering
\small
\renewcommand{\arraystretch}{1.15} 
\begin{tabular}{ccc} 
\hline
\multicolumn{1}{c}{Category}                             & \multicolumn{1}{c}{\begin{tabular}[c]{@{}c@{}}Atomic\textbf{ }\\commands\end{tabular}} & \multicolumn{1}{c}{Example}                                                  \\ 
\hline
\begin{tabular}[c]{@{}c@{}}Text \\Formatting\end{tabular}         & 86                                                                                              & \textbackslash{}textbf                                                                \\ 
\hdashline
\begin{tabular}[c]{@{}c@{}}Equations \\and Symbols\end{tabular}   & 83                                                                                              & \textbackslash{}arcsin                                                                \\ 
\hdashline
\begin{tabular}[c]{@{}c@{}}Document\\Structure\end{tabular}       & 75                                                                                              & \textbackslash{}subsubsection\{\}                                                     \\ 
\hdashline
\begin{tabular}[c]{@{}c@{}}Citation and\\References\end{tabular}  & 39                                                                                              & \begin{tabular}[c]{@{}l@{}}\textbackslash{}bibliographystyle\\\{style\}\end{tabular}  \\ 
\hdashline
\begin{tabular}[c]{@{}c@{}}Tables and\\Figures\end{tabular}       & 36                                                                                              & \textbackslash{}cellcolor\{color\}                                                    \\ 
\hline
Total                                                    & 319                                                                                     &                                                                                       \\
\hline
\end{tabular}
\caption{Details of the atomic LaTeX commands used to build TeXpert}
\label{tab:atomic}
\end{table}

\vspace{0.1cm}
\noindent\textbf{Generating TeXpert using atomic LaTeX commands:} We curate a structured benchmark dataset containing natural language instructions for generating LaTeX code for various elements of scientific content using a combination of manual effort and LLM-based command generation. We build our dataset incrementally (while restricting the domain to atomic commands collected in the previous step to ensure specificity to scientific document components) using three different classes, namely Simple, Average and Hard, by increasing the complexity of tasks, the number of distinct atomic commands and components of scientific documents needed, adding package requirements, and so on. 

In order to classify the final task complexity as Simple, Average or Hard, we use specific constraints based on the number of commands, packages and components, precise description of which, along with a few examples, is found in Table \ref{tab:constraints} in Appendix \ref{sec:appendix-a2}. With a focus on a small but high-quality dataset, we manually verify every row across all three classes in our dataset to ensure clear LaTeX generation requirements and consistency with the difficulty constraints. Our final dataset, named TeXpert, thus contains instructions and a classification label based on difficulty. After experimentation, we also add columns with a LaTeX code satisfying all requirements (if generated by any LLM) for future fine-tuning, along with the LLM that generated this correct code, resulting in the final schema in Figure \ref{fig:process}. Statistics of our dataset are shown in Table \ref{tab:stats}.

\begin{table*}
\centering
\small
\renewcommand{\arraystretch}{1.15} 
\begin{tabular}{ccccc} 
\hline
Difficulty Class & \begin{tabular}[c]{@{}c@{}}No. of\\samples\end{tabular} & \begin{tabular}[c]{@{}c@{}}Average length of\\textual instructions\end{tabular} & \begin{tabular}[c]{@{}c@{}}Average no. of\\atomic LaTeX commands\end{tabular} & \begin{tabular}[c]{@{}c@{}}Average no.\\of extra\\LaTeX packages\end{tabular} \\
\hline
Simple  & 250 & 115.8 ± 24 characters    & 10.9 ± 7.2  & 0.5 ± 0.8  \\
Average & 150 & 299.1 ± 85.7 characters  & 51.2 ± 29.2 & 3.6 ± 2.4  \\
Hard    & 40  & 558.4 ± 216.7 characters & 85.9 ± 31.0 & 6.6 ± 2.0  \\
\hline
\end{tabular}
\caption{Statistics of the TeXpert dataset, organised by difficulty class}
\label{tab:stats}
\end{table*}

\section{Experimental Setup}
We utilise a systematic evaluation framework to assess LLMs' ability to generate syntactically correct LaTeX code from natural language prompts using the TeXpert dataset. We experiment with a wide range of open-source LLMs including Mistral Large 24.11 \citep{mistral2025large2407}, Codestral \citep{mistral2025codestral}, DeepSeek V3 \citep{deepseekai2024deepseekv3technicalreport}, and DeepSeek Coder 33b \citep{guo2024deepseekcoderlargelanguagemodel} as well as multiple high-performance closed-source models including GPT-4o \citep{openai2025gpt4o}, GPT-4o-mini \citep{openai2025gpt4omini}, Gemini 1.5 Flash \citep{geminiteam2024gemini15unlockingmultimodal}, Claude 3.5 Sonnet \citep{anthropic2025claude3} and Grok 2-1212 \citep{xai2025grok2}. 

For each sample across the three difficulty levels in TeXpert, we provide the LLM with a prompt containing task instructions for LaTeX code generation (provided in Figure \ref{fig:prompt-1} in Appendix \ref{sec:appendix-b}). During generation, model parameters were set to pre-determined values to ensure deterministic outputs, as detailed in Table \ref{tab:generation-params} in Appendix \ref{sec:appendix-b}. Detailed model configurations are provided in Section \ref{sec:appendix-b3} in Appendix \ref{sec:appendix-b}. Rule-based extraction techniques are used to extract the LaTeX code from the response. 

We then evaluate each LLM’s response with GPT-4o as a judge, using a predefined evaluation prompt (refer to Figure \ref{fig:prompt-2} in Appendix \ref{sec:appendix-b}) to compute success rates and classify error types (described in Table \ref{tab-err-desc} in Appendix \ref{sec:appendix-b}). The evaluation prompt was iteratively refined through manual spot checks of evaluation outputs, focusing on clarity, correctness, and alignment with evaluation criteria. This process continued until the prompt consistently yielded reliable and interpretable results, as per our judgment. For the hard set, we also provide manually generated and verified LaTeX code as a reference during evaluation, to help identify all requirements of the task. To mitigate potential evaluation bias from using the same model family as the judge, we use DeepSeek v3 as an evaluator for GPT-4o and GPT-4o-mini.

\section{Result Discussion}
\begin{figure}[h]
  \includegraphics[width=\columnwidth]{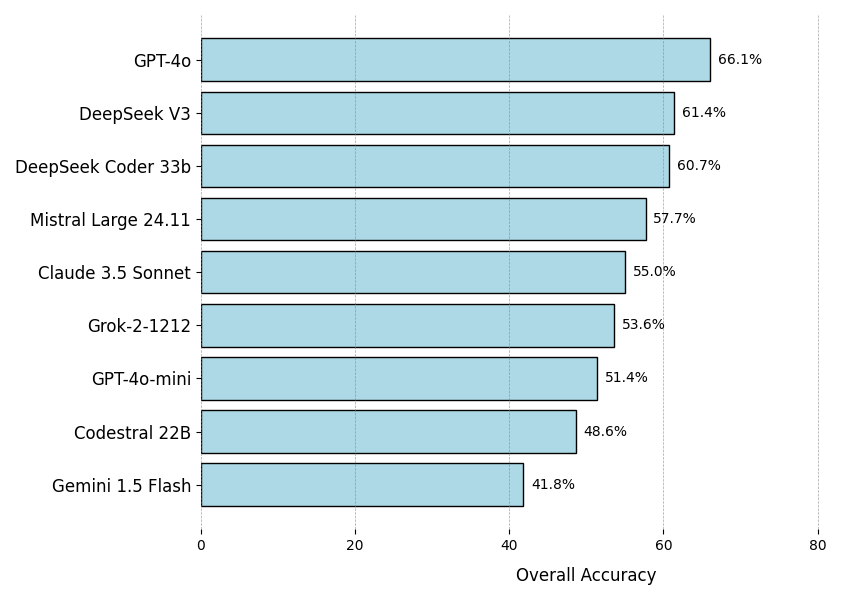}
  \caption{Overall accuracy for LaTeX generation tasks by various LLMs}
  \label{fig:bar}
\end{figure}

\begin{figure}[h]
  \includegraphics[width=\columnwidth]{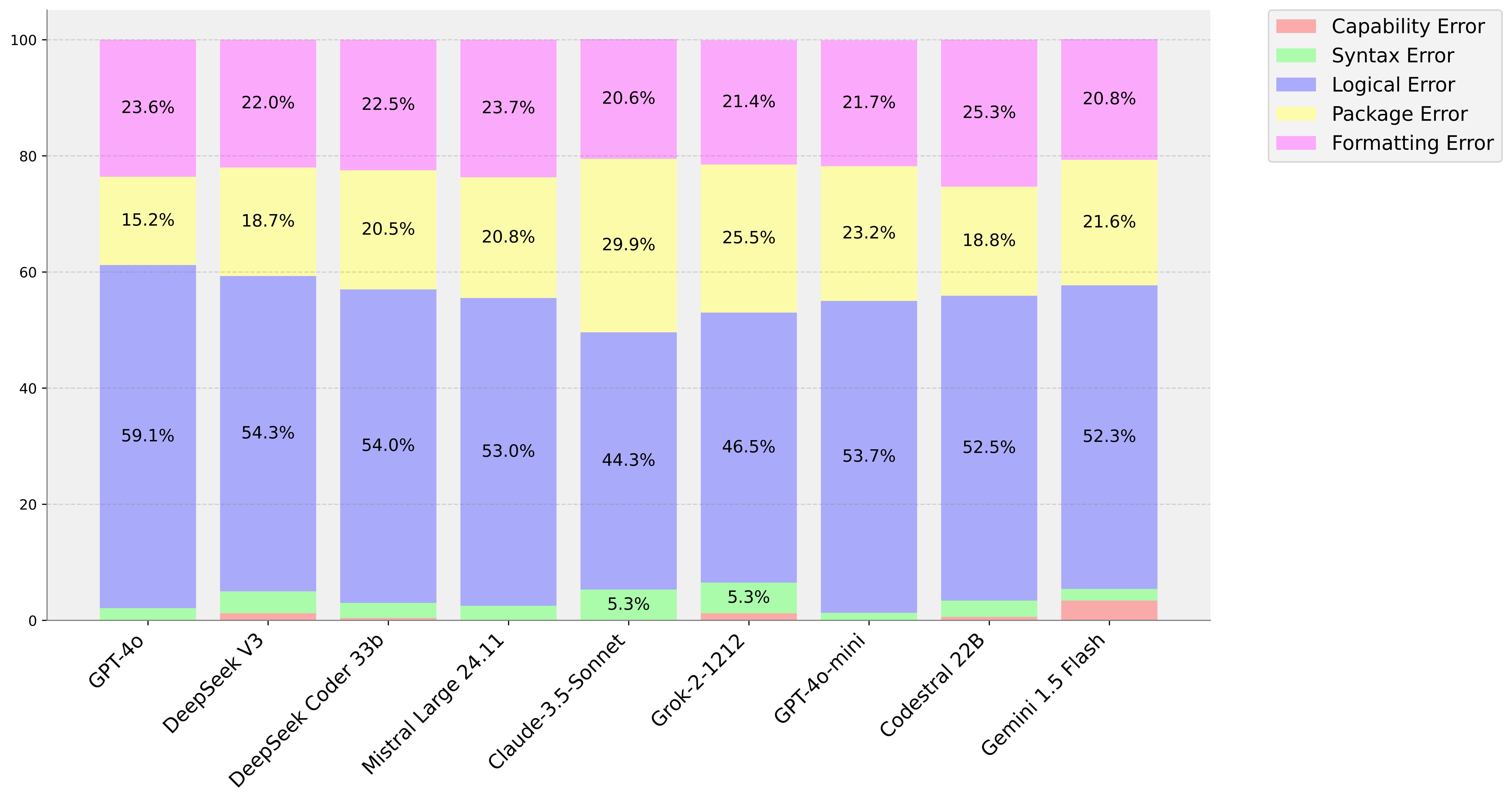}
  \caption{Error distribution for LaTeX generation tasks by various LLMs}
  \label{fig:error}
\end{figure}

The accuracy of LaTeX generation for scientific documents across difficulty classes is presented in Table \ref{tab:results} and visualised in Figure \ref{fig:bar}. The overall distribution of error types across all difficulty levels is presented in Table \ref{tab:error} and Figure \ref{fig:error}, while individual error distributions for Simple, Average, and Hard difficulty classes are also provided in Tables \ref{tab:simple-err}, \ref{tab:avg-error} and \ref{tab:hard-err} in Appendix \ref{sec:appendix-b2}, respectively. From Table \ref{tab:results}, we can infer that GPT-4o outshines all other LLMs in LaTeX code generation, closely followed by DeepSeek v3. DeepSeek Coder 33b provides the best performance on the most complex tasks. 

\begin{table}
\centering
\small
\renewcommand{\arraystretch}{1.12} 
\begin{tabular}{ccccc} 
\hline
\multirow{2}{*}{Model} & \multicolumn{4}{c}{Accuracy \%} \\ 
\cdashline{2-5}
                       & Simple & Average & Hard & Overall \\ 
\hline
\multicolumn{5}{c}{Closed-Source Models} \\ 
\hline
\begin{tabular}[c]{@{}c@{}}GPT-4o-\\mini\end{tabular}         & 62.4           & 45.3           & 5             & 51.4           \\
\hdashline
GPT-4o                                                       & \textbf{78.8}  & 58.7           & 15            & \textbf{66.1}  \\
\hdashline
\begin{tabular}[c]{@{}c@{}}Claude-3.5\\Sonnet\end{tabular}    & 62.8           & 56.7           & 0             & 55.0           \\
\hdashline
\begin{tabular}[c]{@{}c@{}}Gemini 1.5\\Flash\end{tabular}     & 53.6           & 33.3           & 0             & 41.8           \\
\hdashline
\begin{tabular}[c]{@{}c@{}}Grok 2\\1212\end{tabular}          & 62.4           & 52.0           & 5             & 53.6           \\ 
\hline
\multicolumn{5}{c}{Open-Source Models} \\ 
\hline
\begin{tabular}[c]{@{}c@{}}Mistral\\Large 24.11\end{tabular}  & 64.4           & \textbf{59.33} & 10            & 57.7           \\
\hdashline
\begin{tabular}[c]{@{}c@{}}Codestral\\22B\end{tabular}        & 60.8           & 41.3           & 0             & 48.6           \\
\hdashline
\begin{tabular}[c]{@{}c@{}}DeepSeek\\V3\end{tabular}          & 71.2           & 58.7           & 10            & 61.4           \\
\hdashline
\begin{tabular}[c]{@{}c@{}}DeepSeek\\Coder 33b\end{tabular}   & 69.2           & 58.0           & \textbf{17.5} & 60.7           \\
\hline
\end{tabular}
\caption{Main accuracy results (in \%). Values in bold indicate the best accuracy for each difficulty class}
\label{tab:results}
\end{table}

\begin{table}
\centering
\small
\renewcommand{\arraystretch}{1.1} 
\begin{tabular}{cccccc} 
\hline
\multirow{2}{*}{Model} & \multicolumn{5}{c}{Error Types in \%} \\ 
\cdashline{2-6}
                       & CE   & SE   & LE   & PE   & FE   \\ 
\hline
\multicolumn{6}{c}{Closed-Source Models} \\ 
\hline
\begin{tabular}[c]{@{}c@{}}GPT-4o-\\mini\end{tabular}         & 0.0  & 1.3  & 53.7 & 23.2 & 21.7 \\
\hdashline
GPT-4o                                                       & 0.0  & 2.1  & 59.1 & 15.2 & 23.6 \\
\hdashline
\begin{tabular}[c]{@{}c@{}}Claude-3.5\\Sonnet\end{tabular}    & 0.0  & 5.3  & 44.3 & 29.9 & 20.6 \\
\hdashline
\begin{tabular}[c]{@{}c@{}}Gemini 1.5\\Flash\end{tabular}     & 3.4  & 2.0  & 52.3 & 21.6 & 20.8 \\
\hdashline
\begin{tabular}[c]{@{}c@{}}Grok-2\\1212\end{tabular}          & 1.2  & 5.3  & 46.5 & 25.5 & 21.4 \\ 
\hline
\multicolumn{6}{c}{Open-Source Models} \\ 
\hline
\begin{tabular}[c]{@{}c@{}}Mistral\\Large 24.11\end{tabular}  & 0.0  & 2.5  & 53.0 & 20.8 & 23.7 \\
\hdashline
\begin{tabular}[c]{@{}c@{}}Codestral\\22B\end{tabular}        & 0.6  & 2.8  & 52.5 & 18.8 & 25.3 \\
\hdashline
\begin{tabular}[c]{@{}c@{}}DeepSeek\\V3\end{tabular}          & 1.2  & 3.8  & 54.3 & 18.7 & 22.0 \\
\hdashline
\begin{tabular}[c]{@{}c@{}}DeepSeek\\Coder 33b\end{tabular}   & 0.4  & 2.6  & 54.0 & 20.5 & 22.5 \\
\hline
\end{tabular}
\caption{Overall error distribution for LaTeX generation tasks by various LLMs. CE = Capability Error, SE = Syntax Error, LE = Logical Error, PE = Package Error, FE = Formatting Error}
\label{tab:error}
\end{table}

\noindent\textbf{LaTeX generation tasks expose fundamental LLM shortcomings:} Even models that perform highly on other benchmarks like GPT-4o and Mistral Large fail to achieve over 80\% and 60\% accuracy in simple and average sets, respectively. This reveals a critical capability gap in using LLMs for formatting scientific documents in LaTeX, most likely due to the scarcity of LaTeX examples in training datasets.

\vspace{0.1cm}
\noindent\textbf{Hard LaTeX tasks reveal a universal limitation across models:} Accuracy across the Simple and Average sets remains consistent across models, however, models show a dramatic performance cliff on hard tasks, with Claude and Gemini completely failing. This consistent degradation pattern clearly shows a threshold on the number and complexity of instructions for LaTeX generation using LLMs, which can be presumed to lie between instruction statistics for the Average and Hard sets in Table \ref{tab:stats}.

\vspace{0.2cm}
\noindent\textbf{Open-source models strongly rival closed-source ones in LaTeX generation:} Open-source models like DeepSeek V3 and DeepSeek Coder 33b perform well on par with frontier closed-source models like GPT-4o and Claude-3.5-Sonnet in overall accuracy with minimal capability errors as well. Notably, DeepSeek Coder 33b greatly outperforms Claude 3.5 Sonnet and Grok 2 in the Hard set. This demonstrates the potential of open-source models to provide powerful yet cost-effective alternatives.

\section{Error Analysis}
In this section, we provide a brief analysis of the most common error types and probable sources during LaTeX generation by LLMs. From our perspective, most powerful LLMs still struggle to provide error-free code due to basic oversights like missing packages and unfaithful instruction following. It is encouraging to see minimal capability errors and syntax errors. We leave an in-depth analysis of the root cause of errors to the future scope. 

\noindent\textbf{Logical errors dominate:} Logical errors consistently account for the majority of issues across LLMs, highlighting struggles to fully satisfy task requirements. In all the cases we analysed, the most pronounced errors across all model variants were focused on missed instructions and wrong structural placement, especially in GPT-4o-mini and all open-source models. Similarly, error clustering in multiple equation and table generation tasks indicates that LLMs like DeepSeek v3 and Mistral Large struggle with maintaining long-range consistency. We believe these errors likely arise from weak structural understanding inherent in LLMs, limited exposure to LaTeX context, and misalignment between pretraining tasks and formal document generation.

\vspace{0.1cm}
\noindent\textbf{Frequent formatting lapses:} Notably, formatting errors occur far more frequently than we anticipated in all the LLMs we experimented with. Analysis of the evaluations reveals that these errors primarily involve incorrect environment selection and malformed tables or captions accompanying large tables or figures. Such issues indicate limited structural understanding and inadequate grounding in LaTeX syntax, even in larger models like DeepSeek v3 and GPT-4o, showing that scale alone is not the solution. We speculate that these errors stem from a scarcity of training data and examples specifically addressing table formatting and related constructs.

\vspace{0.1cm}
\noindent\textbf{Package errors are concerning:} Package errors are prominently caused by improper or incomplete inclusion and configuration of essential LaTeX packages, especially bibliography-related ones, most prominent in Claude 3.5 Sonnet. GPT-4o has the lowest share of missing packages, showing encouraging signs that more inclusive training data might mitigate this issue, although Codestral's minimal package error rate also suggests potential for alternative approaches to reduce them further. Additionally, the use of non-standard or incompatible packages, especially in DeepSeek and Mistral models, is concerning and may point to LLMs hallucinating or making up packages to fill reasoning gaps. Overall, package issues suggest a fundamental gap in dependency management and environment consistency within LaTeX code generated by LLMs.

\section{Conclusion}
We curate TeXpert, a comprehensive benchmark designed to challenge LLMs to evaluate their LaTeX code generation capability from natural language prompts. Our dataset consists of a total of 440 high-quality samples, organised by difficulty. Our findings reveal that LaTeX generation is still an underperforming skill in LLMs and that there is a need to include LaTeX package details and complex layouts in the training data for LLMs to improve their capability in this task. By making the code and dataset for TeXpert publicly available, we hope to support and encourage further research within the community.

\section*{Limitations}
Our research marks a significant step forward in providing a benchmark for evaluating the LaTeX generation capabilities of LLMs. However, we acknowledge the limitations of our work as follows:
\begin{itemize}[left=0em] 
    \item \textbf{Limited dataset size:} The Hard set's restricted size of 40 samples is a possible challenge in the generalisability of our findings. To address this, we encourage future work to increase the number and complexity of hard examples to broaden the benchmark’s effectiveness.
    \item \textbf{Fine-tuning models and improved prompts:} Using our dataset to fine-tune models and reduce logical and package errors in LaTeX-based tasks is another straightforward extension to our work, along with checking advanced prompting structures for performance improvements.
    \item \textbf{Additional LaTeX sources and applications:} While our work focuses on generating LaTeX code for only scientific documents, incorporating sources and tasks for other document types, such as resumes and books, would broaden the research scope.
\end{itemize}  

\bibliography{acl_latex}

\begin{thebibliography}{27}
\providecommand{\natexlab}[1]{#1}

\bibitem[{Abacha et~al.(2025)Abacha, wai Yim, Fu, Sun, Yetisgen, Xia, and Lin}]{abacha2025medecbenchmarkmedicalerror}
Asma~Ben Abacha, Wen wai Yim, Yujuan Fu, Zhaoyi Sun, Meliha Yetisgen, Fei Xia, and Thomas Lin. 2025.
\newblock \href {https://arxiv.org/abs/2412.19260} {Medec: A benchmark for medical error detection and correction in clinical notes}.
\newblock \emph{Preprint}, arXiv:2412.19260.

\bibitem[{AI(2024{\natexlab{a}})}]{mistral2025codestral}
Mistral AI. 2024{\natexlab{a}}.
\newblock \href {https://mistral.ai/news/codestral/} {Codestral}.
\newblock Accessed: 2025-01-05.

\bibitem[{AI(2024{\natexlab{b}})}]{mistral2025large2407}
Mistral AI. 2024{\natexlab{b}}.
\newblock \href {https://mistral.ai/news/mistral-large-2407/} {Mistral large}.
\newblock Accessed: 2025-01-05.

\bibitem[{Anthropic(2024)}]{anthropic2025claude3}
Anthropic. 2024.
\newblock \href {https://www-cdn.anthropic.com/fed9cc193a14b84131812372d8d5857f8f304c52/Model_Card_Claude_3_Addendum.pdf} {Model card claude 3 addendum}.
\newblock Accessed: 2025-01-05.

\bibitem[{Bos and McCurley(2023)}]{bos2023latexmetadatapublishingworkflows}
Joppe~W. Bos and Kevin~S. McCurley. 2023.
\newblock \href {https://arxiv.org/abs/2301.08277} {Latex, metadata, and publishing workflows}.
\newblock \emph{Preprint}, arXiv:2301.08277.

\bibitem[{Chamezopoulos et~al.(2024)Chamezopoulos, Herrmannova, De~Waard, Herrmannova, Rosati, and Kashnitsky}]{chamezopoulos-etal-2024-overview}
Savvas Chamezopoulos, Drahomira Herrmannova, Anita De~Waard, Drahomira Herrmannova, Domenic Rosati, and Yury Kashnitsky. 2024.
\newblock \href {https://aclanthology.org/2024.sdp-1.2/} {Overview of the {D}ag{P}ap24 shared task on detecting automatically generated scientific paper}.
\newblock In \emph{Proceedings of the Fourth Workshop on Scholarly Document Processing (SDP 2024)}, pages 7--11, Bangkok, Thailand. Association for Computational Linguistics.

\bibitem[{Chen et~al.(2024)Chen, Liao, Qi, Eustratiadis, Monz, Bisazza, and de~Rijke}]{chen-etal-2024-sifo}
Xinyi Chen, Baohao Liao, Jirui Qi, Panagiotis Eustratiadis, Christof Monz, Arianna Bisazza, and Maarten de~Rijke. 2024.
\newblock \href {https://doi.org/10.18653/v1/2024.findings-emnlp.92} {The {SIF}o benchmark: Investigating the sequential instruction following ability of large language models}.
\newblock In \emph{Findings of the Association for Computational Linguistics: EMNLP 2024}, pages 1691--1706, Miami, Florida, USA. Association for Computational Linguistics.

\bibitem[{DeepSeek-AI(2024)}]{deepseekai2024deepseekv3technicalreport}
DeepSeek-AI. 2024.
\newblock \href {https://arxiv.org/abs/2412.19437} {Deepseek-v3 technical report}.
\newblock \emph{Preprint}, arXiv:2412.19437.

\bibitem[{Deng et~al.(2017)Deng, Kanervisto, Ling, and Rush}]{deng2017imagetomarkupgenerationcoarsetofineattention}
Yuntian Deng, Anssi Kanervisto, Jeffrey Ling, and Alexander~M. Rush. 2017.
\newblock \href {https://arxiv.org/abs/1609.04938} {Image-to-markup generation with coarse-to-fine attention}.
\newblock \emph{Preprint}, arXiv:1609.04938.

\bibitem[{Garc{\'i}a-Ferrero et~al.(2024)Garc{\'i}a-Ferrero, Agerri, Atutxa~Salazar, Cabrio, de~la Iglesia, Lavelli, Magnini, Molinet, Ramirez-Romero, Rigau, Villa-Gonzalez, Villata, and Zaninello}]{garcia-ferrero-etal-2024-medmt5}
Iker Garc{\'i}a-Ferrero, Rodrigo Agerri, Aitziber Atutxa~Salazar, Elena Cabrio, Iker de~la Iglesia, Alberto Lavelli, Bernardo Magnini, Benjamin Molinet, Johana Ramirez-Romero, German Rigau, Jose~Maria Villa-Gonzalez, Serena Villata, and Andrea Zaninello. 2024.
\newblock \href {https://aclanthology.org/2024.lrec-main.974/} {{M}ed{MT}5: An open-source multilingual text-to-text {LLM} for the medical domain}.
\newblock In \emph{Proceedings of the 2024 Joint International Conference on Computational Linguistics, Language Resources and Evaluation (LREC-COLING 2024)}, pages 11165--11177, Torino, Italia. ELRA and ICCL.

\bibitem[{Guo et~al.(2024)Guo, Zhu, Yang, Xie, Dong, Zhang, Chen, Bi, Wu, Li, Luo, Xiong, and Liang}]{guo2024deepseekcoderlargelanguagemodel}
Daya Guo, Qihao Zhu, Dejian Yang, Zhenda Xie, Kai Dong, Wentao Zhang, Guanting Chen, Xiao Bi, Y.~Wu, Y.~K. Li, Fuli Luo, Yingfei Xiong, and Wenfeng Liang. 2024.
\newblock \href {https://arxiv.org/abs/2401.14196} {Deepseek-coder: When the large language model meets programming -- the rise of code intelligence}.
\newblock \emph{Preprint}, arXiv:2401.14196.

\bibitem[{He et~al.(2024)He, Zeng, He, Liang, and Xiao}]{he-etal-2024-complex}
Qianyu He, Jie Zeng, Qianxi He, Jiaqing Liang, and Yanghua Xiao. 2024.
\newblock \href {https://doi.org/10.18653/v1/2024.findings-emnlp.637} {From complex to simple: Enhancing multi-constraint complex instruction following ability of large language models}.
\newblock In \emph{Findings of the Association for Computational Linguistics: EMNLP 2024}, pages 10864--10882, Miami, Florida, USA. Association for Computational Linguistics.

\bibitem[{Hendrycks et~al.(2021)Hendrycks, Burns, Kadavath, Arora, Basart, Tang, Song, and Steinhardt}]{hendrycks2021measuringmathematicalproblemsolving}
Dan Hendrycks, Collin Burns, Saurav Kadavath, Akul Arora, Steven Basart, Eric Tang, Dawn Song, and Jacob Steinhardt. 2021.
\newblock \href {https://arxiv.org/abs/2103.03874} {Measuring mathematical problem solving with the math dataset}.
\newblock \emph{Preprint}, arXiv:2103.03874.

\bibitem[{Jung et~al.(2024)Jung, Hyeon, Kwon, Kim, Ryu, Lee, and Do}]{jung2024mathbridgelargecorpusdataset}
Kyudan Jung, Sieun Hyeon, Jeong~Youn Kwon, Nam-Joon Kim, Hyun~Gon Ryu, Hyuk-Jae Lee, and Jaeyoung Do. 2024.
\newblock \href {https://arxiv.org/abs/2408.07081} {Mathbridge: A large corpus dataset for translating spoken mathematical expressions into $latex$ formulas for improved readability}.
\newblock \emph{Preprint}, arXiv:2408.07081.

\bibitem[{LaTeX()}]{latex_project}
LaTeX.
\newblock An introduction to {L}a{T}e{X}.
\newblock \url{https://www.latex-project.org/about/}.
\newblock Accessed: 2025-01-04.

\bibitem[{OpenAI(2024{\natexlab{a}})}]{openai2025gpt4omini}
OpenAI. 2024{\natexlab{a}}.
\newblock \href {https://openai.com/index/gpt-4o-mini-advancing-cost-efficient-intelligence/} {Gpt-4o mini: Advancing cost-efficient intelligence}.
\newblock Accessed: 2025-01-05.

\bibitem[{OpenAI(2024{\natexlab{b}})}]{openai2025gpt4o}
OpenAI. 2024{\natexlab{b}}.
\newblock \href {https://openai.com/index/gpt-4o-system-card/} {Gpt-4o system card}.
\newblock Accessed: 2025-01-05.

\bibitem[{Qin et~al.(2024)Qin, Song, Hu, Yao, Cho, Wang, Wu, Liu, Liu, and Yu}]{qin-etal-2024-infobench}
Yiwei Qin, Kaiqiang Song, Yebowen Hu, Wenlin Yao, Sangwoo Cho, Xiaoyang Wang, Xuansheng Wu, Fei Liu, Pengfei Liu, and Dong Yu. 2024.
\newblock \href {https://doi.org/10.18653/v1/2024.findings-acl.772} {{I}n{F}o{B}ench: Evaluating instruction following ability in large language models}.
\newblock In \emph{Findings of the Association for Computational Linguistics: ACL 2024}, pages 13025--13048, Bangkok, Thailand. Association for Computational Linguistics.

\bibitem[{Roberts et~al.(2024)Roberts, Lee, Wong, Yasunaga, Mai, and Liang}]{roberts2024image2structbenchmarkingstructureextraction}
Josselin~Somerville Roberts, Tony Lee, Chi~Heem Wong, Michihiro Yasunaga, Yifan Mai, and Percy Liang. 2024.
\newblock \href {https://arxiv.org/abs/2410.22456} {Image2struct: Benchmarking structure extraction for vision-language models}.
\newblock \emph{Preprint}, arXiv:2410.22456.

\bibitem[{Sherifi et~al.(2024)Sherifi, Slhoub, and Nembhard}]{sherifi2024potential}
Betim Sherifi, Khaled Slhoub, and Fitzroy Nembhard. 2024.
\newblock \href {https://arxiv.org/abs/2501.00217} {The potential of llms in automating software testing: From generation to reporting}.
\newblock \emph{Preprint}, arXiv:2501.00217.

\bibitem[{Tang et~al.(2024)Tang, Zong, Phang, Zhao, Zhou, Cohan, and Gerstein}]{tang2024strucbenchlargelanguagemodels}
Xiangru Tang, Yiming Zong, Jason Phang, Yilun Zhao, Wangchunshu Zhou, Arman Cohan, and Mark Gerstein. 2024.
\newblock \href {https://arxiv.org/abs/2309.08963} {Struc-bench: Are large language models really good at generating complex structured data?}
\newblock \emph{Preprint}, arXiv:2309.08963.

\bibitem[{Team(2024)}]{geminiteam2024gemini15unlockingmultimodal}
Gemini Team. 2024.
\newblock \href {https://arxiv.org/abs/2403.05530} {Gemini 1.5: Unlocking multimodal understanding across millions of tokens of context}.
\newblock \emph{Preprint}, arXiv:2403.05530.

\bibitem[{xAI(2024)}]{xai2025grok2}
xAI. 2024.
\newblock \href {https://x.ai/blog/grok-2} {Grok 2}.
\newblock Accessed: 2025-01-05.

\bibitem[{Yin et~al.(2023)Yin, Ye, Liu, Ren, and Sch{\"u}tze}]{yin-etal-2023-llm}
Wenpeng Yin, Qinyuan Ye, Pengfei Liu, Xiang Ren, and Hinrich Sch{\"u}tze. 2023.
\newblock \href {https://doi.org/10.18653/v1/2023.emnlp-tutorial.4} {{LLM}-driven instruction following: Progresses and concerns}.
\newblock In \emph{Proceedings of the 2023 Conference on Empirical Methods in Natural Language Processing: Tutorial Abstracts}, pages 19--25, Singapore. Association for Computational Linguistics.

\bibitem[{Zhang et~al.(2024)Zhang, Jiang, Zhang, Lin, Guo, Qiu, Zhou, Lu, Chang, Gao, and Li}]{zhang2024mathversedoesmultimodalllm}
Renrui Zhang, Dongzhi Jiang, Yichi Zhang, Haokun Lin, Ziyu Guo, Pengshuo Qiu, Aojun Zhou, Pan Lu, Kai-Wei Chang, Peng Gao, and Hongsheng Li. 2024.
\newblock \href {https://arxiv.org/abs/2403.14624} {Mathverse: Does your multi-modal llm truly see the diagrams in visual math problems?}
\newblock \emph{Preprint}, arXiv:2403.14624.

\bibitem[{Zhao et~al.(2024)Zhao, Singh, Bhathena, Ramos, Joshi, Gadiyaram, and Sharma}]{zhao-etal-2024-optimizing}
Yiyun Zhao, Prateek Singh, Hanoz Bhathena, Bernardo Ramos, Aviral Joshi, Swaroop Gadiyaram, and Saket Sharma. 2024.
\newblock \href {https://doi.org/10.18653/v1/2024.naacl-industry.23} {Optimizing {LLM} based retrieval augmented generation pipelines in the financial domain}.
\newblock In \emph{Proceedings of the 2024 Conference of the North American Chapter of the Association for Computational Linguistics: Human Language Technologies (Volume 6: Industry Track)}, pages 279--294, Mexico City, Mexico. Association for Computational Linguistics.

\bibitem[{Zou et~al.(2024)Zou, Wang, Thakur, and Kani}]{zou2024stempomevaluatinglanguagemodels}
Jiaru Zou, Qing Wang, Pratyush Thakur, and Nickvash Kani. 2024.
\newblock \href {https://arxiv.org/abs/2411.00387} {Stem-pom: Evaluating language models math-symbol reasoning in document parsing}.
\newblock \emph{Preprint}, arXiv:2411.00387.

\end{thebibliography}

\appendix

\section{Curation of TeXpert - Additional Details}
\label{sec:appendix-a}
\subsection{Data Collection and Sources}
\label{sec:appendix-a1}
To build the core of our TeXpert dataset, we manually extracted atomic commands from the Overleaf documentation listed in row 1 of Table \ref{tab:data-sources} and from 25 documents each in LaTeX template repositories given in rows 2 and 3 of Table \ref{tab:data-sources}. This approach ensured a diverse range of document formats and LaTeX commands commonly used in scientific materials. For each document, a Python script using regular expressions was used to extract atomic LaTeX commands. These commands were then manually verified and grouped into five categories based on their function, as shown in Table \ref{tab:atomic}. This process was intended to focus the dataset on commonly used LaTeX elements in scientific writing.

\begin{table*}[h]
\small
\centering
\renewcommand{\arraystretch}{1.3} 
\begin{tabular}{p{0.35\linewidth} p{0.6\linewidth}}
\hline
\multicolumn{1}{c}{Data Source} & \multicolumn{1}{c}{URL} \\
\hline
Overleaf Documentation & \url{https://www.overleaf.com/learn} \\
Overleaf Academic Journal Templates & \url{https://www.overleaf.com/latex/templates/tagged/academic-journal} \\
LaTeX Templates (Creodocs) & \url{https://www.latextemplates.com/cat/academic-journals} \\
\hline
\end{tabular}
\caption{Primary sources used for collecting atomic LaTeX commands}
\label{tab:data-sources}
\end{table*}

\subsection{Difficulty constraints}
\label{sec:appendix-a2}
Table \ref{tab:constraints} shows the constraints followed while classifying samples into difficulty classes (Simple/Average/Hard) during the generation of tasks in the TeXpert dataset. A randomly chosen example from each set is also provided for reference. 

\begin{table*}
\centering
\small
\renewcommand{\arraystretch}{1.2} 
\begin{tabular}{cccccc} 
\hline
\begin{tabular}[c]{@{}c@{}}Difficulty\\Class\end{tabular} & 
\begin{tabular}[c]{@{}c@{}}Length of\\textual\\instructions\end{tabular} & 
\begin{tabular}[c]{@{}c@{}}No. of \\atomic LaTeX\\Commands\end{tabular} & 
\begin{tabular}[c]{@{}c@{}}No. of\\extra LaTeX \\packages\end{tabular} & 
\begin{tabular}[c]{@{}c@{}}No. of specific\\formatting \\instructions \\(for tables, \\figures, etc.)\end{tabular} & 
Example \\ 
\hline
Simple & 
\begin{tabular}[c]{@{}c@{}}<200\\characters\end{tabular} & 
10–20 & 
<2 & 
<2 & 
\begin{tabular}[c]{@{}c@{}}Create a document with centered \\text in one block and justified text \\in another block.\end{tabular} \\ 
\hdashline
Average & 
\begin{tabular}[c]{@{}c@{}}200–500\\characters\end{tabular} & 
12–80 & 
2–5 & 
2–5 & 
\begin{tabular}[c]{@{}c@{}}Create a document with two sections. \\The first section should contain an \\aligned set of equations. The second \\section should contain a centered table,\\and the table should reference a figure \\placed in the first section.\end{tabular} \\ 
\hdashline
Hard & 
\begin{tabular}[c]{@{}c@{}}500+\\characters\end{tabular} & 
80+ & 
5+ & 
5+ & 
\begin{tabular}[c]{@{}c@{}}Your task is to produce a scientific \\research paper for arXiv that has a title\\page with author names, abstract and \\keywords, table of contents, and several \\sections. Add a 3x3 table that has lists \\in the second column, and figures with \\bold captions in last column. On every \\page except the first, add a footer with \\a signature image. Add an appendix that \\includes a table with header row entirely \\merged. Finally, add a custom \\bibliography.\end{tabular} \\
\hline
\end{tabular}
\caption{Description of constraints used during classification of tasks in TeXpert with a few examples}
\label{tab:constraints}
\end{table*}

\section{Experimentation - Additional Details}
\label{sec:appendix-b}
\subsection{Prompts}
\label{sec:appendix-b1}
\begin{figure}[h]
\centering
  \includegraphics[width=0.81\columnwidth]{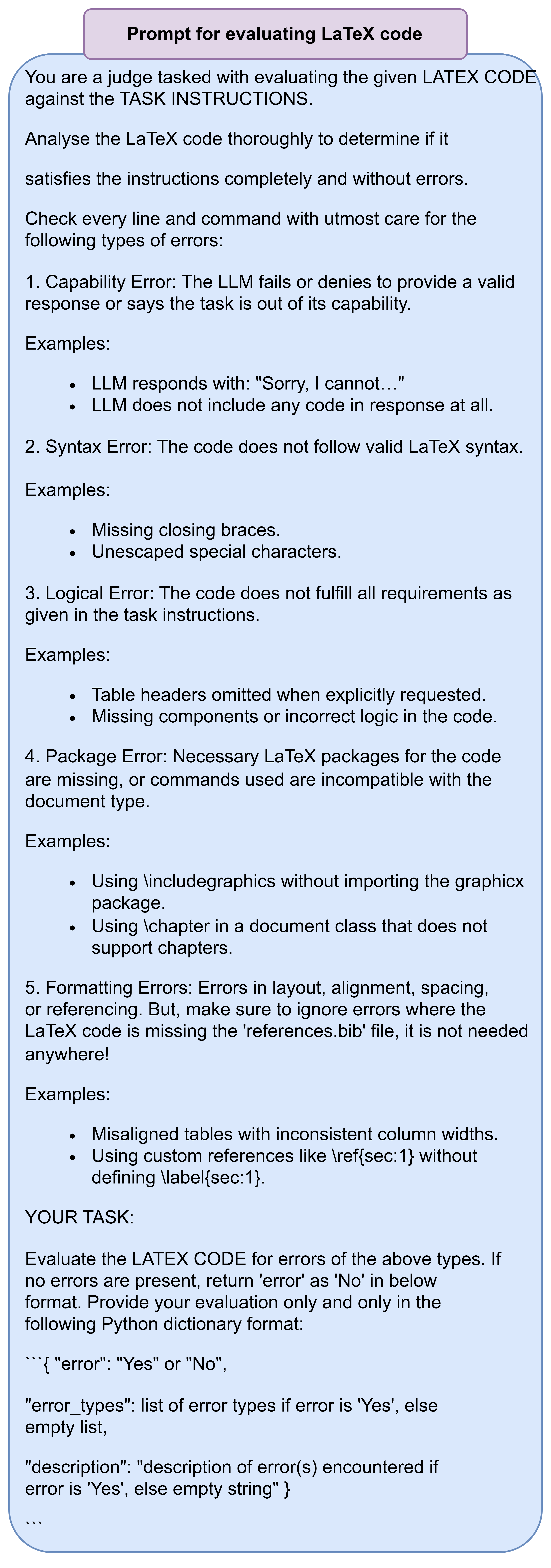}
  \caption{System prompt used to evaluate LaTeX code generated by LLMs using GPT-4o/DeepSeek v3 as-a-judge}
  \label{fig:prompt-2}
\end{figure}

The prompts used during experimentation to evaluate responses using GPT-4o/DeepSeek v3 as a judge and to generate LaTeX code using natural language instructions and are given in Figures \ref{fig:prompt-2} and \ref{fig:prompt-2} respectively.

\subsection{Error descriptions and distribution}
\label{sec:appendix-b2}
Details of error types along with examples are given in Table \ref{tab-err-desc}. Additionally, the individual error distributions for Simple, Average, and Hard difficulty classes for each LLM are given in Tables \ref{tab:simple-err}, \ref{tab:avg-error} and \ref{tab:hard-err} respectively.

\subsection{Model parameters}
\label{sec:appendix-b3}
We report the generation parameters for all models used in our experiments to ensure transparency and reproducibility. All models were accessed through provider APIs, and the common parameter settings used across all models (except Anthropic models) are listed in Table \ref{tab:generation-params}. The model sizes of all closed-source models are approximate and taken from \citet{abacha2025medecbenchmarkmedicalerror}.

\noindent\textbf{OpenAI Models:} We run our experiments on two flagship models, GPT-4o (\textasciitilde200B parameters) and GPT-40-mini (\textasciitilde8B parameters). We use the OpenAI Python SDK to access the models via API, specifying \verb|seed=1234| and \verb|n=1| along with the parameter values listed in Table \ref{tab:generation-params}, to ensure maximum determinism in responses. All other parameters are kept to default values.

\noindent\textbf{DeepSeek Models:} We use two recently released models, DeepSeek v3 (\textasciitilde671B parameters) and DeepSeek Coder (\textasciitilde33B parameters). DeepSeek models were accessed using the OpenAI Python SDK by specifying the DeepSeek URL endpoint and authentication details. Here too, we set \verb|seed=1234| and \verb|n=1| along with the parameter values listed in Table \ref{tab:generation-params} during experimentation, keeping the rest to default values.

\noindent\textbf{Mistral Models:} We experiment with two powerful models, Mistral-Large-Instruct-2411 (\textasciitilde123B parameters) and Codestral-22B-v0.1 (\textasciitilde22B parameters). Both models were accessed using the official API in Mistral Python SDK, with an extra parameter \verb|random_seed=1234| along with values in Table \ref{tab:generation-params}, with the rest as default.

\noindent\textbf{Google AI Models:} The Gemini 1.5 flash model was accessed using the official Google Generative AI Python SDK. Within the Generation Config, we set parameters values to those mentioned in Table \ref{tab:generation-params}, along with \verb|candidate_count=1| and the rest as default.

\noindent\textbf{xAI Models:} We use a recently released Grok-2-1212 model by xAI, accessed using the OpenAI Python SDK by specifying the xAI endpoint. Here too, we set \verb|seed=1234| and \verb|n=1| along with the parameter values listed in Table \ref{tab:generation-params} during experimentation, keeping the rest to default values.

\noindent\textbf{Anthropic Models:} The Claude 3.5 Sonnet model (\textasciitilde175B parameters) was accessed via the official Anthropic Python SDK. Due to limited configurable parameters, only \verb|temperature=0.0|, \verb|top_p=1|, and \verb|max_tokens=8096| were explicitly set, with all other settings left at their defaults.

\begin{table*}
\centering
\small
\renewcommand{\arraystretch}{1.2} 
\begin{tabular}{ccc} 
\hline
Error Type & 
Description & 
Examples \\
\hline
\begin{tabular}[c]{@{}c@{}}Capability\\Error\end{tabular} & 
\begin{tabular}[c]{@{}c@{}}The LLM fails or denies to provide a \\valid response or says the task is out \\of its capability.\end{tabular} & 
\begin{tabular}[c]{@{}c@{}}• LLM responds with: "Sorry, I cannot…"\\• LLM does not include any code in response\end{tabular} \\
\hdashline
\begin{tabular}[c]{@{}c@{}}Syntax\\Error\end{tabular} & 
\begin{tabular}[c]{@{}c@{}}The code generated by the LLM does \\not follow valid LaTeX syntax.\end{tabular} & 
\begin{tabular}[c]{@{}c@{}}• Missing closing braces\\• Unescaped special characters\end{tabular} \\
\hdashline
\begin{tabular}[c]{@{}c@{}}Logical\\Error\end{tabular} & 
\begin{tabular}[c]{@{}c@{}}Mismatches between user instructions \\and the code logic, i.e., requirements \\given in natural language are not \\satisfied by the LaTeX code.\end{tabular} & 
\begin{tabular}[c]{@{}c@{}}• Table headers omitted when explicitly requested\\• Missing components in code\end{tabular} \\
\hdashline
\begin{tabular}[c]{@{}c@{}}Package\\Error\end{tabular} & 
\begin{tabular}[c]{@{}c@{}}Required LaTeX packages are missing \\or commands do not match the \\document type.\end{tabular} & 
\begin{tabular}[c]{@{}c@{}}• Using \texttt{\textbackslash includegraphics} without \\importing the \texttt{graphicx} package\end{tabular} \\
\hdashline
\begin{tabular}[c]{@{}c@{}}Formatting\\\& Referencing\\Error\end{tabular} & 
\begin{tabular}[c]{@{}c@{}}Layout issues like inconsistent alignment, \\font size, or spacing; improper formatting \\for cross-references, citations, or labels.\end{tabular} & 
\begin{tabular}[c]{@{}c@{}}• Misaligned tables with inconsistent widths\\• Using \texttt{\textbackslash ref\{sec:1\}} without defining \\ \texttt{\textbackslash label\{sec:1\}}\end{tabular} \\
\hline
\end{tabular}
\caption{Description and examples of error types used during evaluation of generated LaTeX code by LLMs}
\label{tab-err-desc}
\end{table*}

\begin{figure}[h]
  \includegraphics[width=\columnwidth]{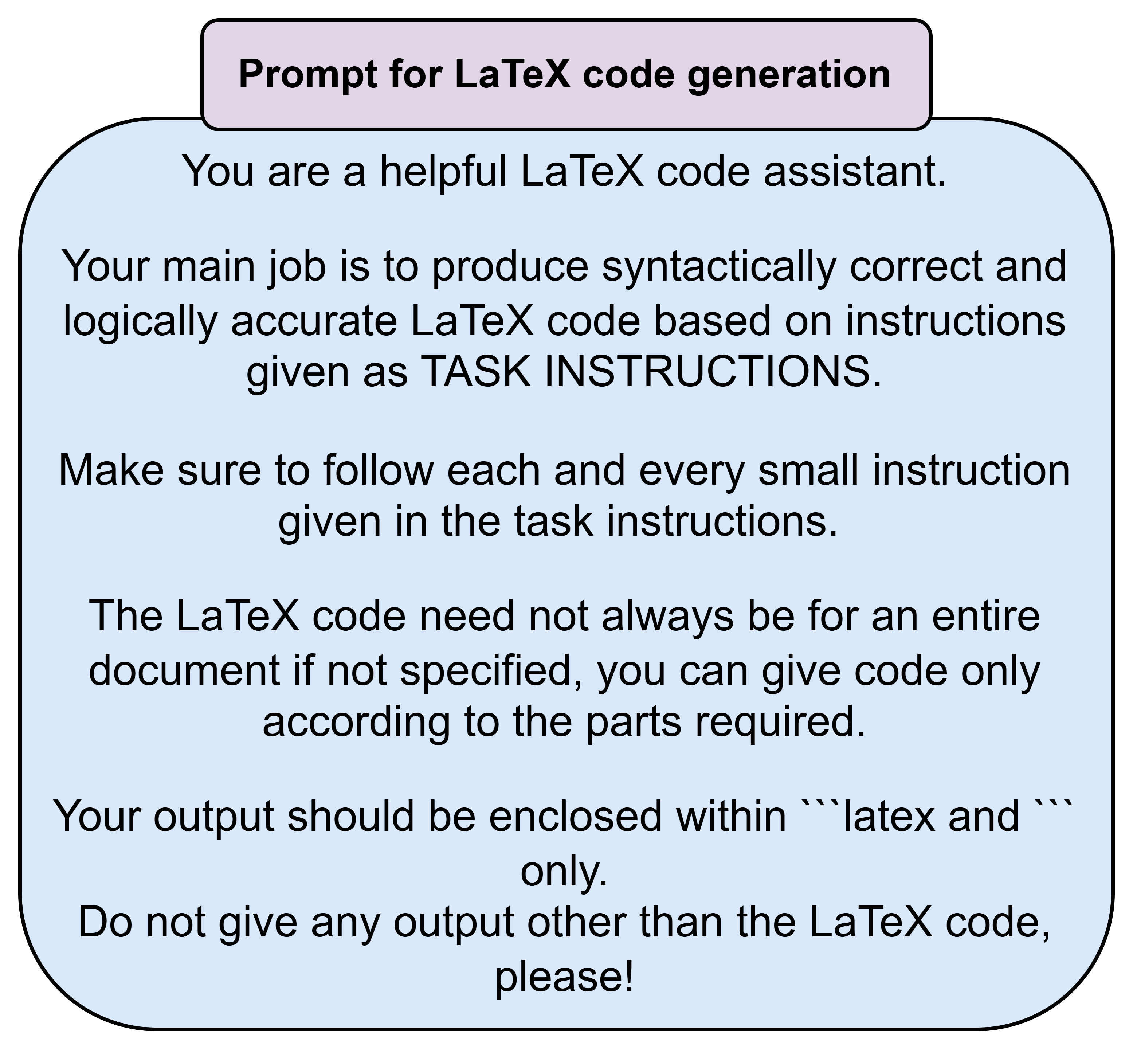}
  \caption{System prompt used to generate LaTeX code using LLMs for given textual instructions}
  \label{fig:prompt-1}
\end{figure}

\begin{table}
\small
\centering
\renewcommand{\arraystretch}{1.1} 
\begin{tabular}{llllll} 
\hline
\multicolumn{1}{c}{\multirow{2}{*}{Model}}                    & \multicolumn{5}{c}{Error
  Types in \% - Simple Set}                                                                                  \\ 
\cline{2-6}
\multicolumn{1}{c}{}                                          & \multicolumn{1}{c}{CE} & \multicolumn{1}{c}{SE} & \multicolumn{1}{c}{LE} & \multicolumn{1}{c}{PE} & \multicolumn{1}{c}{FE}  \\ 
\hline
\multicolumn{6}{c}{Closed-Source Models}                                                                                                                                                    \\ 
\hline
\begin{tabular}[c]{@{}l@{}}GPT-4o\\mini\end{tabular}          & 0                      & 0                      & 57                     & 37.4                   & 5.6                     \\
GPT-4o                                                        & 0                      & 6.3                    & 66.67                  & 17.5                   & 9.5                     \\
\begin{tabular}[c]{@{}l@{}}Claude-3.5\\Sonnet\end{tabular}    & 0                      & 13.6                   & 36.8                   & 41.6                   & 8                       \\
\begin{tabular}[c]{@{}l@{}}Gemini 1.5\\~Flash\end{tabular}    & 5.7                    & 2.9                    & 55                     & 27.9                   & 8.6                     \\
\begin{tabular}[c]{@{}l@{}}Grok-2\\1212\end{tabular}          & 3.6                    & 9.9                    & 43.2                   & 41.4                   & 1.8                     \\ 
\hline
\multicolumn{6}{c}{Open-Source Models}                                                                                                                                                      \\ 
\hline
\begin{tabular}[c]{@{}l@{}}Mistral \\Large 24.11\end{tabular} & 0                      & 7.6                    & 55.5                   & 28.6                   & 8.4                     \\
\begin{tabular}[c]{@{}l@{}}Codestral \\22B\end{tabular}       & 1.7                    & 5                      & 60.3                   & 22.3                   & 10.7                    \\
\begin{tabular}[c]{@{}l@{}}DeepSeek \\V3\end{tabular}         & 3.5                    & 5.9                    & 52.9                   & 30.6                   & 7.1                     \\
\begin{tabular}[c]{@{}l@{}}DeepSeek \\Coder 33b\end{tabular}  & 0                      & 2.2                    & 54.4                   & 36.7                   & 6.7                     \\
\hline
\end{tabular}
\caption{Error distribution for LaTeX generation tasks from the Simple set by various LLMs. CE = Capability Error, SE = Syntax Error, LE = Logical Error, PE = Package Error, FE = Formatting Error}
\label{tab:simple-err}
\end{table}

\begin{table}
\small
\centering
\renewcommand{\arraystretch}{1.1} 
\begin{tabular}{llllll} 
\hline
\multicolumn{1}{c}{\multirow{2}{*}{Model}}                    & \multicolumn{5}{c}{Error
  Types in \% - Average Set}                                                                                  \\ 
\cline{2-6}
\multicolumn{1}{c}{}                                          & \multicolumn{1}{c}{CE} & \multicolumn{1}{c}{SE} & \multicolumn{1}{c}{LE} & \multicolumn{1}{c}{PE} & \multicolumn{1}{c}{FE}  \\ 
\hline
\multicolumn{6}{c}{Closed-Source Models}                                                                                                                                                    \\ 
\hline
\begin{tabular}[c]{@{}l@{}}GPT-4o\\mini\end{tabular}          & 0                      & 1.6                    & 55.9                   & 11.8                   & 30.7                    \\
GPT-4o                                                        & 0                      & 0                      & 58.3                   & 15.7                   & 26                      \\
\begin{tabular}[c]{@{}l@{}}Claude-3.5\\Sonnet\end{tabular}    & 0                      & 1                      & 48.5                   & 24.8                   & 25.7                    \\
\begin{tabular}[c]{@{}l@{}}Gemini 1.5\\~Flash\end{tabular}    & 4.6                    & 0.7                    & 53.6                   & 16.3                   & 24.8                    \\
\begin{tabular}[c]{@{}l@{}}Grok-2\\1212\end{tabular}          & 0                      & 3.6                    & 54.5                   & 17.4                   & 24.5                    \\ 
\hline
\multicolumn{6}{c}{Open-Source Models}                                                                                                                                                      \\ 
\hline
\begin{tabular}[c]{@{}l@{}}Mistral \\Large 24.11\end{tabular} & 0                      & 0                      & 54.1                   & 17.3                   & 28.6                    \\
\begin{tabular}[c]{@{}l@{}}Codestral \\22B\end{tabular}       & 0                      & 1.3                    & 53.9                   & 16.2                   & 28.6                    \\
\begin{tabular}[c]{@{}l@{}}DeepSeek \\V3\end{tabular}         & 0                      & 2.2                    & 60                     & 11.1                   & 26.7                    \\
\begin{tabular}[c]{@{}l@{}}DeepSeek \\Coder 33b\end{tabular}  & 1.1                    & 4.2                    & 56.8                   & 7.4                    & 30.5                    \\
\hline
\end{tabular}
\caption{Error distribution for LaTeX generation tasks from the Average set by various LLMs. CE = Capability Error, SE = Syntax Error, LE = Logical Error, PE = Package Error, FE = Formatting Error}
\label{tab:avg-error}
\end{table}

\begin{table}
\small 
\centering
\renewcommand{\arraystretch}{1.1} 
\begin{tabular}{llllll} 
\hline
\multicolumn{1}{c}{\multirow{2}{*}{Model}}                    & \multicolumn{5}{c}{Error
  Types in \% - Hard Set}                                                                                  \\ 
\cline{2-6}
\multicolumn{1}{c}{}                                          & \multicolumn{1}{c}{CE} & \multicolumn{1}{c}{SE} & \multicolumn{1}{c}{LE} & \multicolumn{1}{c}{PE} & \multicolumn{1}{c}{FE}  \\ 
\hline
\multicolumn{6}{c}{Closed-Source Models}                                                                                                                                                    \\ 
\hline
\begin{tabular}[c]{@{}l@{}}GPT-4o\\mini\end{tabular}          & 0                      & 2.4                    & 48.2                   & 20.5                   & 28.9                    \\
GPT-4o                                                        & 0                      & 0                      & 52.3                   & 12.3                   & 35.4                    \\
\begin{tabular}[c]{@{}l@{}}Claude-3.5\\Sonnet\end{tabular}    & 0                      & 1.2                    & 47.6                   & 23.2                   & 28                      \\
\begin{tabular}[c]{@{}l@{}}Gemini 1.5\\~Flash\end{tabular}    & 0                      & 2.4                    & 48.2                   & 20.5                   & 28.9                    \\
\begin{tabular}[c]{@{}l@{}}Grok-2\\1212\end{tabular}          & 0                      & 2.5                    & 41.8                   & 17.7                   & 38                      \\ 
\hline
\multicolumn{6}{c}{Open-Source Models}                                                                                                                                                      \\ 
\hline
\begin{tabular}[c]{@{}l@{}}Mistral \\Large 24.11\end{tabular} & 0                      & 0                      & 49.3                   & 16.4                   & 34.2                    \\
\begin{tabular}[c]{@{}l@{}}Codestral \\22B\end{tabular}       & 0                      & 2.2                    & 43.3                   & 17.8                   & 36.7                    \\
\begin{tabular}[c]{@{}l@{}}DeepSeek \\V3\end{tabular}         & 0                      & 3.2                    & 50                     & 14.5                   & 32.3                    \\
\begin{tabular}[c]{@{}l@{}}DeepSeek \\Coder 33b\end{tabular}  & 0                      & 1.4                    & 50.7                   & 17.4                   & 30.4                    \\
\hline
\end{tabular}
\caption{Error distribution for LaTeX generation tasks from the Hard set by various LLMs. CE = Capability Error, SE = Syntax Error, LE = Logical Error, PE = Package Error, FE = Formatting Error}
\label{tab:hard-err}
\end{table}

\begin{table}[h]
\small
\centering
\renewcommand{\arraystretch}{1.3} 
\begin{tabular}{p{0.4\linewidth} p{0.4\linewidth}}
\hline
\multicolumn{2}{c}{Generation Parameters} \\
\hline
\texttt{temperature} & \texttt{0.0} \\
\texttt{top\_p} & \texttt{1.0} \\
\texttt{max\_tokens} & \texttt{8096} \\
\texttt{frequency\_penalty} & \texttt{0.0} \\
\texttt{presence\_penalty} & \texttt{0.0} \\
\hline
\end{tabular}
\caption{Generation parameters used across all models}
\label{tab:generation-params}
\end{table}

\end{document}